\documentclass[conference]{ieeeconf}
\IEEEoverridecommandlockouts

\usepackage[backend=biber,
            hyperref=true,
            url=false,
            isbn=false,
            doi=false,
            backref=false,
            style=ieee,
            citestyle=numeric-comp,
            sorting=nyt,
            block=none]{biblatex}
        
\renewcommand{\bibfont}{\small}
\usepackage{flushend}
\usepackage{balance}

\usepackage{marginnote}
\usepackage{graphics}
\usepackage[pdftex]{graphicx}
\usepackage{wrapfig}
\DeclareGraphicsExtensions{.pdf,.png,.jpg}

\usepackage[font={small}]{caption}
\usepackage{subcaption}
\usepackage[rightcaption]{sidecap}
\usepackage{pbox}

\usepackage{mathtools}
\usepackage{amsmath, amssymb, amscd}
\usepackage{ wasysym } 
\usepackage{amsfonts}
\usepackage{mathptmx} 
\usepackage{gensymb} 
\usepackage{nicefrac}       

\DeclareMathAlphabet{\mathcal}{OMS}{lmsy}{m}{n}
\DeclareSymbolFont{largesymbols}{OMX}{cmex}{m}{n}
\usepackage{textcomp} 

\usepackage{array} 
\usepackage{tabularx}
\usepackage{multirow}
\usepackage{multicol}
\usepackage{booktabs}

\usepackage[T1]{fontenc} 
\usepackage[utf8]{inputenc}
\usepackage[english]{babel} 
\usepackage{units}
\usepackage{bm}
\usepackage{times} 
\usepackage{xspace}
\usepackage{flushend}
\usepackage{csquotes}
\usepackage{makeidx}

\let\labelindent\relax
\usepackage{enumitem}

\usepackage{soul} 
\usepackage{subfiles} 

\usepackage[protrusion=true,expansion=true]{microtype}
\usepackage[yyyymmdd]{datetime}

\date{\protect\formatdate{1}{1}{2001}}

\usepackage{url}
\makeatletter
\g@addto@macro{\UrlBreaks}{\UrlOrds}
\makeatother
\usepackage{color}
\usepackage[usenames,dvipsnames,table]{xcolor}
\usepackage[pdfborder={0 0 0.5}]{hyperref}
\hypersetup{
    colorlinks=true,
    linkcolor=black,
    citecolor=black,
    filecolor=cyan,
    urlcolor=black
}

\usepackage{soul} 

\newcommand{\tocite}[1]{%
\textcolor{red}{[cite:\ifthenelse{\equal{#1}{}}{}{#1}?]}
}

\newcommand{\ignore}[1]{}

\newcommand{\figref}[1]{Figure\,\ref{fig:#1}}

\newcommand{\tabref}[1]{Table~\ref{tab:#1}}

\usepackage{algorithm}
\usepackage{algpseudocode}

\newcommand{\algoName}{NTP\xspace}
\newcommand{\algoFull}{Neural Task Programming\xspace}

\begin{document}

\title{Neural Task Programming: \\Learning to Generalize Across Hierarchical Tasks \vspace{-10pt}}

\author{%
Danfei Xu$^{*1}$,
Suraj Nair$^{*2}$,
Yuke Zhu$^{1}$,
Julian Gao$^{1}$, 
Animesh Garg$^{1}$,
Li Fei-Fei$^{1}$,
Silvio Savarese$^{1}$
\thanks{\vspace{-10pt} \hrule \vspace{2pt} * These authors contributed equally to the paper}
\thanks{ $^{1}\,$Stanford Vision \& Learning Lab, $^{2}\,$CS, Caltech.
}%
}

\maketitle

\begin{abstract}
In this work, we propose a novel robot learning framework called Neural Task Programming (\algoName), which bridges the idea of few-shot learning from demonstration and neural program induction. \algoName takes as input a task specification (e.g., video demonstration of a task) and recursively decomposes it into finer sub-task specifications. These specifications are fed to a hierarchical neural program, where bottom-level programs are callable subroutines that interact with the environment. We validate our method in three robot manipulation tasks. 
\algoName achieves strong generalization across sequential tasks that exhibit hierarchal and compositional structures.
The experimental results show that \algoName learns to generalize well towards unseen tasks with increasing lengths, variable topologies, and changing objectives.
\href{https://stanfordvl.github.io/ntp/}{\texttt{stanfordvl.github.io/ntp/}}
\end{abstract}

\IEEEpeerreviewmaketitle

\section{Introduction}
\label{sec:intro}

Autonomy in complex manipulation tasks, such as object sorting, assembly, and de-cluttering, requires sequential decision making with prolonged interactions between the robot and the environment. 
Planning for a complex task and, vitally, adapting to new task objectives and initial conditions is a long-standing challenge in robotics~\cite{fikes1972learning, brooks1986robust}.

\begin{figure}[!t]
\centering
  \includegraphics[width=\linewidth]{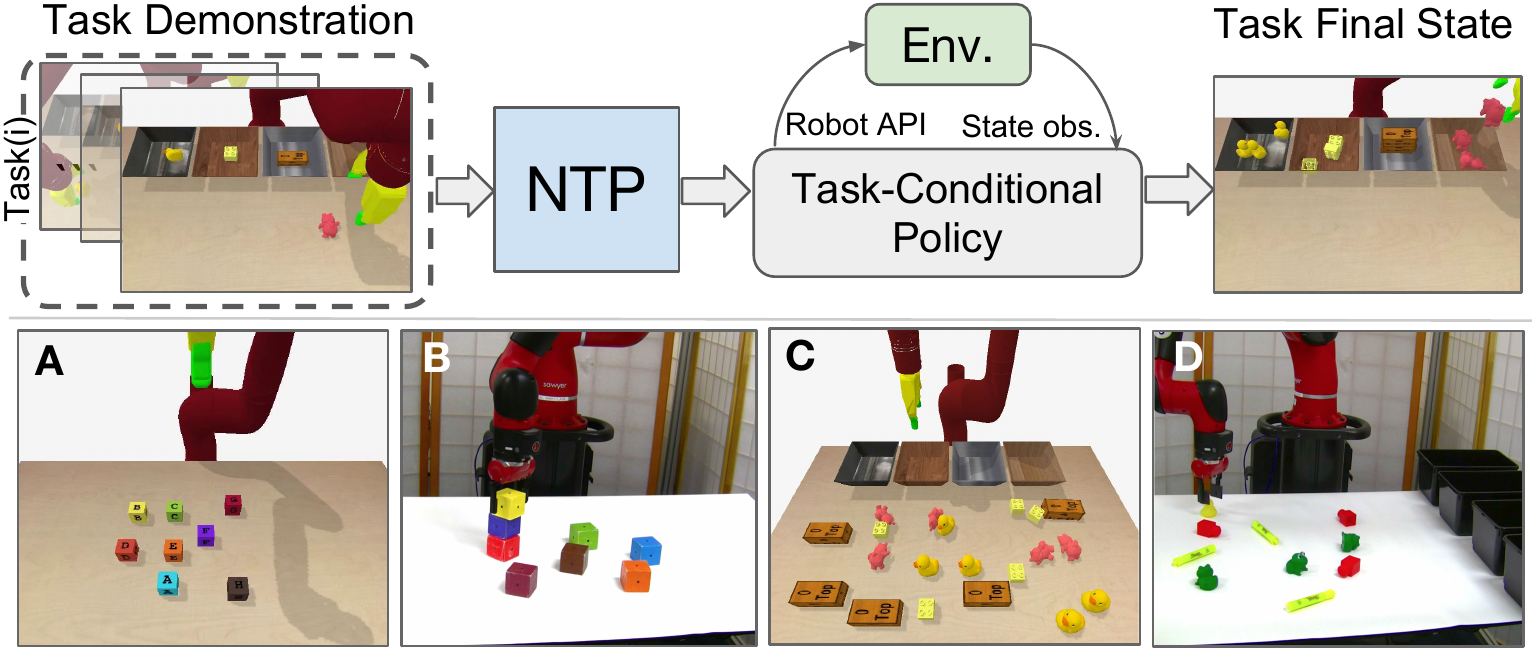}
  \caption{(top) At test time, \algoName instantiates a task-conditional policy (a neural program) that performs the specified task by interpreting a demonstration of a task. The policy interacts with the environment through robot APIs. (bottom) We evaluate \algoName on Block Stacking (A,B), Object Sorting (C, D) and Table Clean-up (\figref{cleanup}) tasks in both simulated and real environment.}
  \label{fig:pull}
  \vspace{-20pt}
\end{figure}

Consider an object sorting task in a warehouse setting -- it requires sorting, retrieval from storage, and packing for shipment.
Each task is a sequence of primitives -- such as \texttt{pick\_up}, \texttt{move\_to}, and \texttt{drop\_into} -- that can be composed into manipulation sub-tasks such as grasping and placing. 
This problem has an expansive space of variations -- different objects-bin maps in sorting, permutations of sub-tasks, varying length order lists -- resulting in a large space of tasks. As a concrete example, \figref{pull}(C) shows a simplified setup of the object sorting task. The task is to transport objects of four categories to four shipping containers. There is a total of 256 possible mappings between object categories and containers, and the variable number of object instances further increases the complexity.   In this paper, we attempt to address two challenges in complex task planning domains, namely (a) \emph{learning policies that generalize to new task objectives}, and (b) \emph{hierarchical composition of primitives for long-term environment interactions}.

We propose \algoFull (\algoName), a unified, task-agnostic learning algorithm that can be applied to a variety of tasks with latent hierarchical structure. 
The key underlying idea is to learn reusable representations shared across tasks and domains.  \algoName interprets a \textit{task specification} (\figref{pull}\,left) and instantiates a hierarchical policy as a neural program (\figref{pull}\,middle), where the bottom-level programs are primitive actions that are executable in the environment. A task specification is defined as a time-series that describes the procedure and the final objective of a task. It can either be a task demonstration recorded as a state trajectory, first/third-person video demonstrations, or even a list of language instructions. In this work, we use \textit{task demonstration} as the task specification. We experiment with two forms of task demonstration: location trajectories of objects that are involved in a task, and a third-person video demonstration of a task. 
\algoName decodes the objective of a task from the input specification and factorizes it into sub-tasks, interacting with the environment with closed-loop feedback until the goal is achieved (\figref{pull}\,right). Each program call takes as input the environment observation and a task specification, producing the next sub-program and a corresponding sub-task specification. The lowest level of the hierarchy is symbolic actions captured through a Robot API. 

This hierarchical decomposition encourages information hiding and modularization, as lower-level modules only access their corresponding sub-task specifications that pertain to their functionality. It prevents the model from learning spurious dependencies on training data, resulting in better reusability.
Essentially, \algoName addresses the key challenges in task generalization: meta-learning for cross-task transfer and hierarchical model to scale to more complex tasks. Hence, \algoName builds on the strengths of neural programming and hierarchical RL while compensating for their shortcomings.

\begin{figure*}[!t]
\centering
\begin{minipage}[t]{.7\linewidth}
    \centering
    \vspace{-5pt}
    \includegraphics[width=0.8\linewidth]{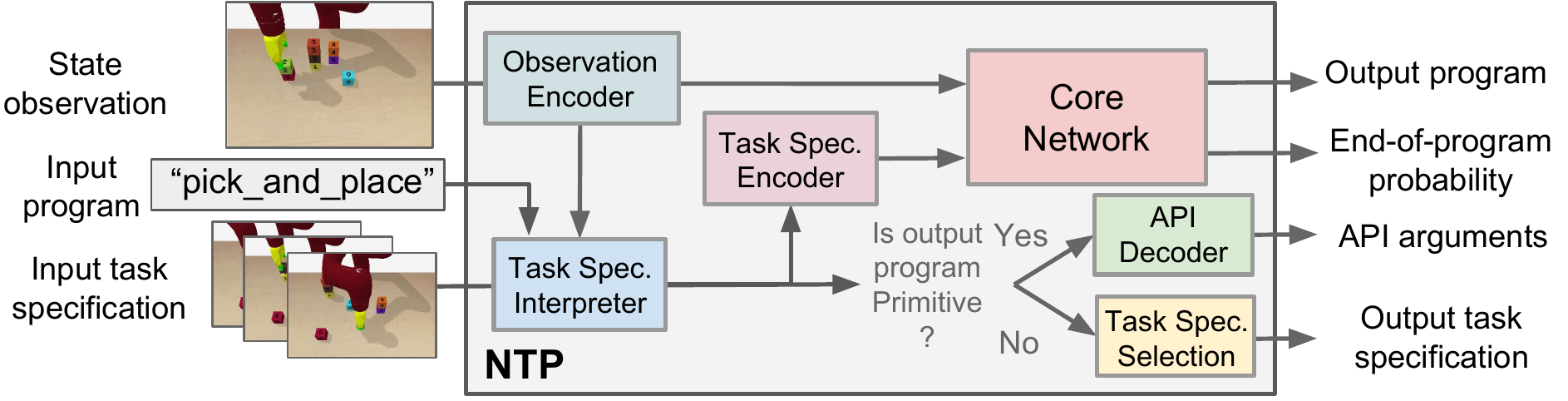}
    \vspace{-5pt}
\end{minipage}%
\begin{minipage}[t]{.295\linewidth}
    \centering
    \caption{\textbf{\algoFull (\algoName)}: Given an input program, a task specification, and the current environment observation, a NTP model predicts the sub-level program to run, the sub-sequence of the task specification that the sub-level program should take as input, and if the current program should stop.}
    \label{fig:model}
    \vspace{-5pt}
\end{minipage}%
\vspace{-20pt}
\end{figure*}

We demonstrate that \algoName generalizes to three kinds of variations in task structure: 1) \emph{Task Length}: varying number of steps due to the increasing problem size (e.g., having more objects to transport); 2) \emph{Task Topology}: the flexible permutations and combinations of sub-tasks to reach the same end goal (e.g., manipulating objects in different orders); and 3) \emph{Task Semantics}: the varying task definitions and success conditions (e.g., placing objects into a different container). 

We evaluate \algoName on three table-top manipulation tasks that require long-term interactions: Block Stacking, Object Sorting, and Table Clean-up. We evaluate each task in both simulated and real-robot setups.

\noindent \textbf{Summary of Contributions:}
\begin{enumerate}[
    topsep=0pt,
    noitemsep,
    leftmargin=*,
    itemindent=12pt]
\item Our primary contribution is a novel modeling framework: \algoName that enables meta-learning on hierarchical tasks. 
\item We show that \algoName enables knowledge transfer and one-shot demonstration based generalization to novel tasks with increasing lengths, varying topology, and changing semantics without restriction on initial configurations.
\item We also demonstrate that \algoName can be trained with visual input (images and video) end-to-end.
\end{enumerate}

\section{Background \& Related work}
\begin{figure*}[!t]
\begin{center}
\includegraphics[width=\linewidth]{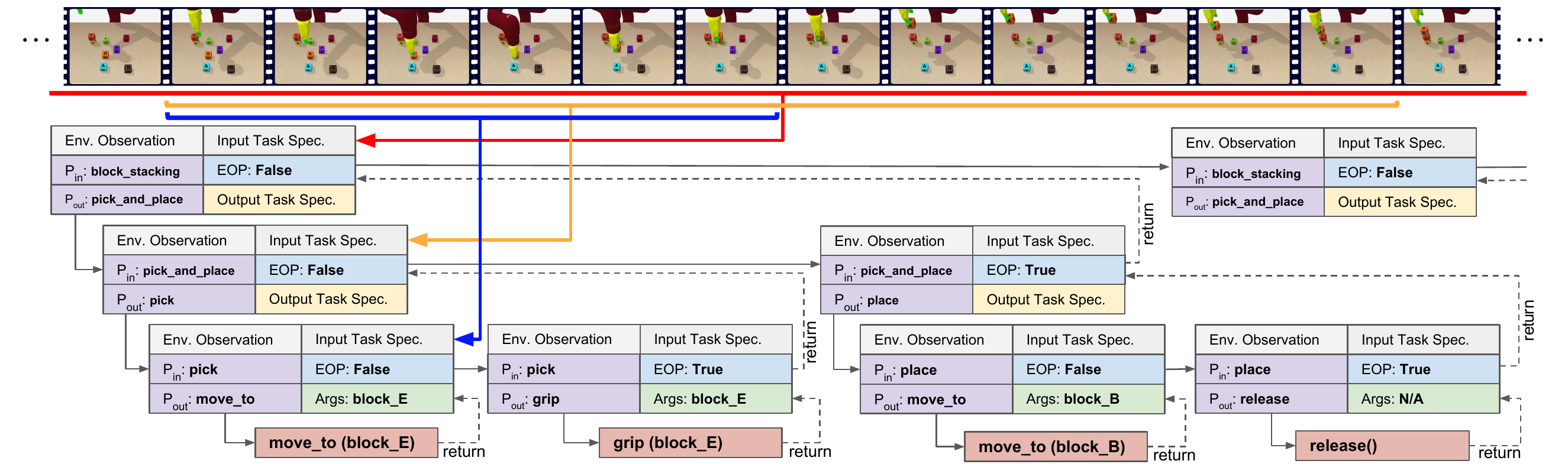}
\caption{
Sample execution trace of \algoName on a block stacking task. The task is to stack lettered blocks into a specified configuration (\texttt{block\_D} on top of \texttt{block\_E}, \texttt{block\_B} on top of \texttt{block\_D}, etc). Top-level program \texttt{block\_stacking} takes in the entire demonstration as input (red window), and predicts the next sub-program to run is \texttt{pick\_and\_place}, and it should take the part of task specification marked by the orange window as the input specification. The bottom-level API call moves the robot and close / open the gripper. When End of Program (EOP) is True, the current program stops and return its caller program.} 
\label{fig:execution}
\vspace{-25pt}
\end{center}
\end{figure*}
\noindent \textbf{Skill Learning}: The first challenge is learning policies that adapt to new task objectives. 
For learning a single task policy, traditional methods often segment a complex task into hand-engineered state machine composed of motion primitives~\cite{fikes1972learning, brooks1986robust,sen2016suturing}. 
Although the model-based approaches are well-founded in principle, they require meticulous model specification and task-specific treatment leading to challenges in scaling. Contrarily, learning-based methods such as reinforcement learning (RL) have yielded promising results using end-to-end policy learning that obviates the need for manually designed state representations through data-driven task-salient features~\cite{mnih2015human,zhu2017icra}. Yet these methods fall short because they need task-specific reward functions~\cite{DBLP:conf/icml/NgHR99}.  

\noindent \textbf{Learning from Demonstrations}: 
LfD fills these gaps by avoiding the need to define state machines or reward functions. The objective in LfD is to learn policies that generalize beyond the provided examples and are robust to perturbations~\cite{argall2009survey,kober2013reinforcement}
A common treatment to LfD is to model data as samples from an expert policy for a fixed task, and use behavior cloning~\cite{jochem1993maniac,ross2011reduction} or reward function approximation~\cite{ng2000algorithms} to output an expert-like policy for that task. 
However, learning policies that generalize to new objectives with LfD remains largely an unexplored problem.

\noindent \textbf{Few-Shot Generalization in LfD}: 
Our work is an instantiation of the decades-old idea of meta-learning with few examples~\cite{fei2006one,vilalta2002perspective}. It has seen a recent revival in deep learning in part because it can address the problems above~\cite{vinyals2016matching}.
Our setting resembles learning by demonstration (LfD) in robotics~\cite{billard2008robot}, particularly one-shot imitation~\cite{wu2010towards,duan2017one}. Our method \emph{learns to learn} from an input task specification during training. At test time, it generates a policy conditioned on a \textit{single} demonstration provided as a time-series showing the task execution. While similar in these aspects, existing works in both skill learning and LfD are inept at tasks with sparse reward functions and complex hierarchical structures such as Montezuma's Revenge~\cite{kulkarni2016hierarchical}.

\noindent \textbf{Hierarchical Skill Composition}: The second challenge we consider is the hierarchical composition of primitives to enable long-term robot-environment interaction. The idea of using hierarchical models for complex tasks has been widely explored in both reinforcement learning and robotics\,\cite{kober2013reinforcement,sung2013learning}. 
A common treatment to manage task structure complexity is to impose hierarchy onto the learned policy. The \emph{options} framework composes primitive actions into multi-step actions, which facilitates policy learning at higher-level semantic and/or temporal abstraction~\cite{fox2017multi,sutton1999between}. 
Notable examples include structured reinforcement learning methods, especially hierarchical variants of RL that handle decomposition through multi-stage policies operating over options~\cite{parr1998reinforcement,BaconHP16option,kulkarni2016hierarchical}.
However, the naïve use of a hierarchical RL model with "sub-policies" or options optimized for a specific task doesn't guarantee modularity or reusability across task objectives.

The core idea of \algoName resonates with recent works on dynamic neural networks, which aim to learn and reuse primitive network modules. These methods have been successfully applied to several domains such as robot control~\cite{andreas2016modular} and visual question answering~\cite{andreas2016neural}. However, they have exhibited limited generalization ability across tasks. In contrast, we approach the problem of hierarchical task learning via neural programming to attain modularization and reusability\,\cite{reed2016neural}. As a result, our model achieves significantly better generalization results than non-hierarchical models such as~\cite{duan2017one}.

\noindent \textbf{FSMs and Neural Program Induction}:
An exciting and non-intuitive insight of this paper is that the well-studied Finite State Machine (FSM) model lends itself to learning reusable hierarchical policies thereby addressing the problem of composability without the need for hand-crafting state transitions.
There have been a few studies learning FSMs from data~\cite{krishnan2018transition,giles2008learning}.
In line with the idea, recent works in neural programming using deep models have enabled symbolic reasoning systems to be trained end-to-end, which have shown potential to handle multi-modal and raw input/out data~\cite{reed2016neural,devlin2017robustfill} and achieve symbolic generalization~\cite{cai2017making}.

\algoName belongs to a family of neural program induction methods, where the goal is to learn a latent program representation that generates program outputs~\cite{reed2016neural,devlin2017robustfill}.
While these models have been shown to generalize on task length, they are tested on basic computational tasks only with limited generalization to task semantics and topology. Similar to NTP, Neural Programmer-Interpreter (NPI)~\cite{reed2016neural} has proposed to use a task-agnostic recurrent neural network to represent and execute programs.
In contrast to previous work on neural program induction, NPI-based models are trained with richer supervision from the full program execution traces and can learn semantically meaningful programs with high data efficiency. However, program induction, including NPI, is not capable of generalizing to novel programs without training.

\algoName is a meta-learning algorithm that learns to instantiate  \textit{neural programs} given demonstrations of tasks, thereby generalizing to unseen tasks/programs. 
Intuitively, \algoName decomposes the overall objective (e.g., object sorting) into simpler objectives (e.g., pick and place) recursively. For each of such sub-tasks, \algoName delegates a neural program to perform the task. 
The neural programs, together with the task decomposition mechanism, are trained end-to-end. 

While previous work has largely focused on executing a pre-defined task one at a time \algoName not only exhibits one-shot generalization to tasks with longer lengths as NPI, but also generalizes to sub-task permutations (topology) and success conditions (semantics).

\section{Problem Formulation}
We consider the problem of an agent performing actions to interact with an environment to accomplish tasks. Let $\mathbb{T}$ be the set of all tasks, $\mathbb{S}$ be the environment state space, and $\mathbb{A}$ be the action space. For each task $t\in \mathbb{T}$, the Boolean function $g: \mathbb{S}\times\mathbb{T} \rightarrow \{0,1\}$ defines the success condition of the task. Given any state $s\in\mathbb{S}$, $g(s, t)=1$ if the task $t$ is completed in the state $s$, and $g(s, t)=0$ otherwise. 
The task space $\mathbb{T}$ can be infinite. We thus need a versatile way to describe the task semantics. We describe each task using a task specification $\psi(t)\in\Psi$, where $\Psi$ is the set of all possible task specifications. Formally, we consider a task specification as a sequence of random variables $\psi(t)=\{x_1, x_2, \ldots, x_N\}$. 

\algoName takes a \textit{task specification} $\psi(t)$ as input in order to instantiate a policy. $\psi(t)$ is defined as a time series that describes the procedure and the final objective of the task. In experiments, we consider two forms of task specifications: trajectories of object locations and raw video sequences.  In many real-world tasks, the agent has no access to the underlying environment states. It only receives a sample of environment observation $o\in\mathbb{O}$ that corresponds to the state $s$, where $\mathbb{O}$ is the observation space. Our goal is to learn a ``meta-policy'' that instantiates a feedback policy from a specification of a task, $\tilde{\pi}: \Psi \rightarrow(\mathbb{O}\rightarrow\mathbb{A})$. At test time, a specification of a new task $\psi(t)$ is input to \algoName. The meta-policy then generates a policy $\pi(a|o; \psi(t)):\mathbb{O}\rightarrow\mathbb{A}$, to reach task-completion state $s_T$ where $g(s_T, t)=1$. 

\noindent \textbf{Why use Neural Programming for LfD?}
Previous work has mostly used a monolithic network architecture to represent a goal-driven policy~\cite{duan2017one,schaul2015universal,zhu2017icra}. These methods cannot exploit the compositional task structures to facilitate modularization and reusability. Instead, we represent our policy $\tilde{\pi}$ as a neural program that takes a task specification as its input argument. As illustrated in \figref{model}, \algoName uses a task-agnostic core network to decide which sub-program to run next and adaptively feeds a subset of the task specification to the next program. 
Intuitively, \algoName recursively decomposes a task specification and solves a hierarchical task by divide-and-conquer. ~\figref{execution} highlights this feature with a sample execution of a task. 
Our method extends upon a special type of neural programming architecture named Neural Programmer-Interpreter (NPI)~\cite{cai2017making,reed2016neural}. NPI generalizes well to input size but cannot generalize to unseen task objectives. \algoName combines the idea of meta-learning and NPI. The ability to interpret task specifications and instantiate policies accordingly makes \algoName generalize across tasks.

\subsection{Neural Programmer-Interpreter (NPI)} 
Before introducing our \algoName model, it is useful to briefly overview the NPI paradigm~\cite{reed2016neural}. NPI is a type of neural program induction algorithm, in which a network is trained to imitate the behavior of a computer program, i.e., the network learns to invoke programs recursively given certain context or stop the current program and return to upper-level programs. 
The core of NPI is a long-short memory (LSTM)~\cite{hochreiter1997long} network. At the $i$-th time step, it selects the next program to run conditioned on the current observation $o_i$ and the previous LSTM hidden units $h_{i-1}$. A domain-specific encoder is used to encode the observation $o_i$ into a state representation $s_i$. The NPI controller takes as input the state $s_i$, the program embedding $p_i$ retrieved from a learnable key-value memory structure [$M^{key};M^{prog}]$, and the current arguments $a_i$. It generates a program key, which is used to invoke a sub-program $p_{i+1}$ using content-based addressing, the arguments to the next program $a_{i+1}$, and the end-of-program probability $r_i$. The NPI model maintains a program call stack. Each time a sub-program is called, the caller's LSTM hidden units embedding and its program embedding is pushed to the stack. Formally, the NPI core has three learnable components, a domain-specific encoder $f_{enc}$, an LSTM $f_{lstm}$, and an output decoder $f_{dec}$. The full update being: $s_i = f_{enc}(o_i, a_i)\quad h_i = f_{lstm}(s_i, p_i, h_{i-1})\quad r_i, p_{i+1}, a_{i+1} = f_{dec}(h_i)$.
When executing a program with the NPI controller, it performs one of the following three things: 1) when the end-of-program probability exceeds a threshold $\alpha$ (set to 0.5), this program is popped up from the stack and control is returned to the called; 2) when the program is not primitive, a sub-program with its arguments is called; and 3) when the program is primitive, a low-level basic action is performed in the environments. The LSTM core is shared across all tasks. 

\vspace{-5pt}
\section{\algoFull}
\label{sec:model}
\noindent \textbf{Overview.} 
\algoName has three key components: Task Specification Interpreter $f_{TSI}$, Task Specification Encoder $f_{TSE}$, and a core network $f_{CN}$ (\figref{model}. The Task Specification Encoder transforms a task specification $\psi_i$ into a vector space. 
The core network takes as input the state $s_i$, the program $p_i$, and the task specification $\psi_i$, producing the next sub-program to invoke $p_{t+1}$ and an end-of-program probability $r_t$. The program returns to the caller when $r_t$ exceeds a threshold $\alpha$ (set to 0.5). We detail the inference procedure in Algorithm~\ref{algo:inference}.

\noindent \textbf{NTP vs NPI}: We highlight three main differences of \algoName than the original NPI: (1) \algoName can interpret task specifications and perform hierarchical decomposition and thus can be considered as a meta-policy; (2) it uses APIs as the primitive actions to scale up neural programs for complex tasks; and (3) it uses a reactive core network instead of a recurrent network, making the model less history-dependent, enabling feedback control for recovery from failures. 
In addition to the three key components, \algoName implements two modules similar to the NPI architectures~\cite{cai2017making,reed2016neural}: (1) domain-specific task encoders that map an observation to a state representation $s_i=f_{ENC}(o_i)$, and (2) a key-value memory that stores and retrieves embeddings: $j^{*}=\arg\max_{j=1\ldots N}(M^{key}_{j,:}k_i)$ and $p_i = M^{prog}_{j^*,:}$, where $k_i$ is the program key predicted by the core network.

\noindent \textbf{Scaling up \algoName with APIs.} The bottom-level programs in NPI correspond to primitive actions that are executable in the environment. To scale up neural programs in coping with the complexity of real-world tasks, it is desirable to use existing tools and subroutines (i.e., motion planner) such that learning can be done at an abstract level. In computer programming, application programming interfaces (APIs) have been a standard protocol for developing software by using basic modules. 
Here we introduce the concept of API to neural programming, where the bottom-level programs correspond to a set of robot APIs, e.g., moving the robot arm using inverse kinematics. 
Each API takes specific arguments, e.g., an object category or the end effector's target position. \algoName jointly learns to select APIs functions and to generate their input arguments. The APIs that are used in the experiments are \texttt{move\_to}, \texttt{grip}, and \texttt{release}. \texttt{move\_to} takes an object index as the API argument and calls external functions to move the gripper to above the object whose position is either given by the simulator or predicted by an object detector. \texttt{grip} closes the gripper and \texttt{release} opens the gripper.

\begin{algorithm}[t!]
\caption{\algoName Inference Procedure}
\label{algo:inference}
\begin{algorithmic}
\State \textbf{Inputs:} task specification $\psi$, program id $i$, and environment observation $o$ 
\Function{RUN}{$i$, $\psi$}
\State $r\gets 0$, $p\gets M^{prog}_{i,:}$, $s\gets f_{ENC}(o)$, $c\gets f_{TSE}(\psi)$
\While {$r < \alpha$}
\State $k, r \gets f_{CN}(c, p, s)$, $\psi_2\gets f_{TSI}(\psi, p, s)$
\State $i_2\gets \arg\max_{j=1\ldots N}(M^{key}_{j,:}k)$
\If {program $i_2$ is primitive} \Comment if $i_2$ is an API
\State $\mathbf{a}\gets f_{TSI}(\psi_2, i_2, s)$ \Comment decode API args
\State \textsc{RUN\_API}($i_2, \mathbf{a}$) \Comment run API $i_2$ with args $\mathbf{a}$
\Else
\State RUN($i_2,\psi_2$) \Comment run program $i_2$ w/ task spec $\psi_2$
\EndIf
\EndWhile
\EndFunction
\end{algorithmic}
\end{algorithm}
\setlength{\textfloatsep}{5pt}

\noindent \textbf{Task Specification Interpreter.} The Task Specification Interpreter, taking a task specification as input, chooses to perform one of the two operations: (1) when the current program $p$ is not primitive, it predicts the sub-task specification for the next sub-program; and (2) when $p$ is primitive (i.e., an API), it predicts the arguments of the API. 

Let $\psi_i$ be the task specification of the $i$-th program call, where $\psi_i$ is a sequence of random variables $\psi_i=\{x_1, x_2, \ldots, x_{N_i}\}$. The next task specification $\psi_{i+1}$ is determined by three inputs: the environment state $s_i$, the current program $p_i$, and the current specification $\psi_i$. When $p_i$ is a primitive, TSI uses an API-specific decoder (i.e., an MLP) to predict the API arguments from the tuple $(s_i, p_i, \psi_i)$. 

We focus on the cases when $p_i$ is not primitive. In this case, TSI needs to predict a sub-task specification $\psi_{i+1}$ for the next program $p_{i+1}$. This sub-task specification should only access relevant information to the sub-task. To encourage information hiding from high-level to low-level programs, we enforce the scoping constraint, such that $\psi_{i+1}$ is a contiguous subsequence of $\psi_i$. Formally, given $\psi_i=\{x_1, x_2, \ldots, x_{N_i}\}$, the goal is to find the optimal contiguous subsequence $\psi_{i+1}=\{x_{p}, x_{p+1}, \ldots, x_{q-1}, x_q\}$, where $1\leq p\leq q\leq N_i$.

\noindent \textbf{Subsequence Selection (Scoping).}
We use a convolutional architecture to tackle the subsequence selection problem. First, we embed each input element $\psi_i=\{x_1, x_2, \ldots, x_{N_i}\}$ into a vector space $\phi_i=\{w_1, w_2, \ldots, w_{N_i}\}$, where each $w_i\in\mathbb{R}^d$. We perform temporal convolution at every temporal location $j$ of the sequence $\phi_i$, where each convolutional kernel is parameterized by $W\in\mathbb{R}^{m\times dk}$ and $b\in\mathbb{R}^{m}$, which takes a concatenation of $k$ consecutive input elements and produces a single output $y^{j}_i\in\mathbb{R}^m$. We use \emph{relu} as the nonlinearities. The outputs from all convolutional kernels $\mathbf{y}^j_i$ are concatenated with the program embedding $p_i$ and the encoded states $s_i$ into a single vector $h_j=[p_i; \mathbf{y}^j_i; s_i]$. Finally, we compute the softmax probability of four scoping labels $\Pr_j(l\in\{\texttt{Start}, \texttt{End}, \texttt{Inside}, \texttt{Outside}\})$. These scoping labels indicate whether this temporal location is the start/end point of the correct subsequence, or if it resides inside/outside the subsequence. 
We use these probabilities to decode the optimal subsequence as the output sub-task specification $\psi_{i+1}$. 

The decoding process can be formulated as the maximum contiguous subsequence sum problem, which can be solved optimally in linear time. However in practice, taking the start and end points with the highest probabilities results in a good performance. In our experiments, we set $\psi_{i+1} = \{x_\text{st}, x_{\text{st}+1}, \ldots, x_\text{ed}\}$, where $st = \arg\max_{j=1\ldots N_i}\Pr_j(\texttt{Start})$ and $ed = \arg\max_{j=1\ldots N_i}\Pr_j(\texttt{End})$. 
This process is illustrated in \figref{execution}, wherein the model factorizes a video sequence which illustrates the procedure of \emph{pick\_and\_place} into a fraction that only illustrates \emph{pick}.
This convolutional TSI architecture is invoked recursively along the program execution trace. It decomposes a long task specification into increasingly fine-grained pieces from high-level to low-level tasks. This method naturally enforces the scoping constraint. Our experimental results show that such information hiding mechanism is crucial to good generalization.

\noindent \textbf{Model Training.} We train the model using rich supervision from program execution traces. Each execution trace is a list of tuples $\{\xi_t\,|\,\xi_t=(\psi_t, p_t, s_t), t=1\ldots T\}$, where $T$ is the length of the execution trace. Our training objective is to maximize the probability of the correct executions over all the tasks in the dataset $\mathcal{D}=\{(\xi_t,\xi_{t+1})\}$, such that
$\theta^* = \Sigma_{\mathcal{D}}\log\Pr[\xi_{t+1}|\xi_t; \theta]$.

We collect a dataset that consists of execution traces from multiple types of tasks and their task specifications.
For each specification, we provide the ground-truth hierarchical decomposition of the specification for training by rolling a hard-coded expert policy. We use cross-entropy loss at every temporal location of the task specification to supervise the scoping labels. We also adopted the idea of adaptive curriculum from NPI~\cite{reed2016neural}, where the frequency of each mini-batch being fetched is proportional to the model's prediction error with respect to the corresponding program.

 \begin{figure}[!t]
\centering
  \includegraphics[width=0.9\linewidth]{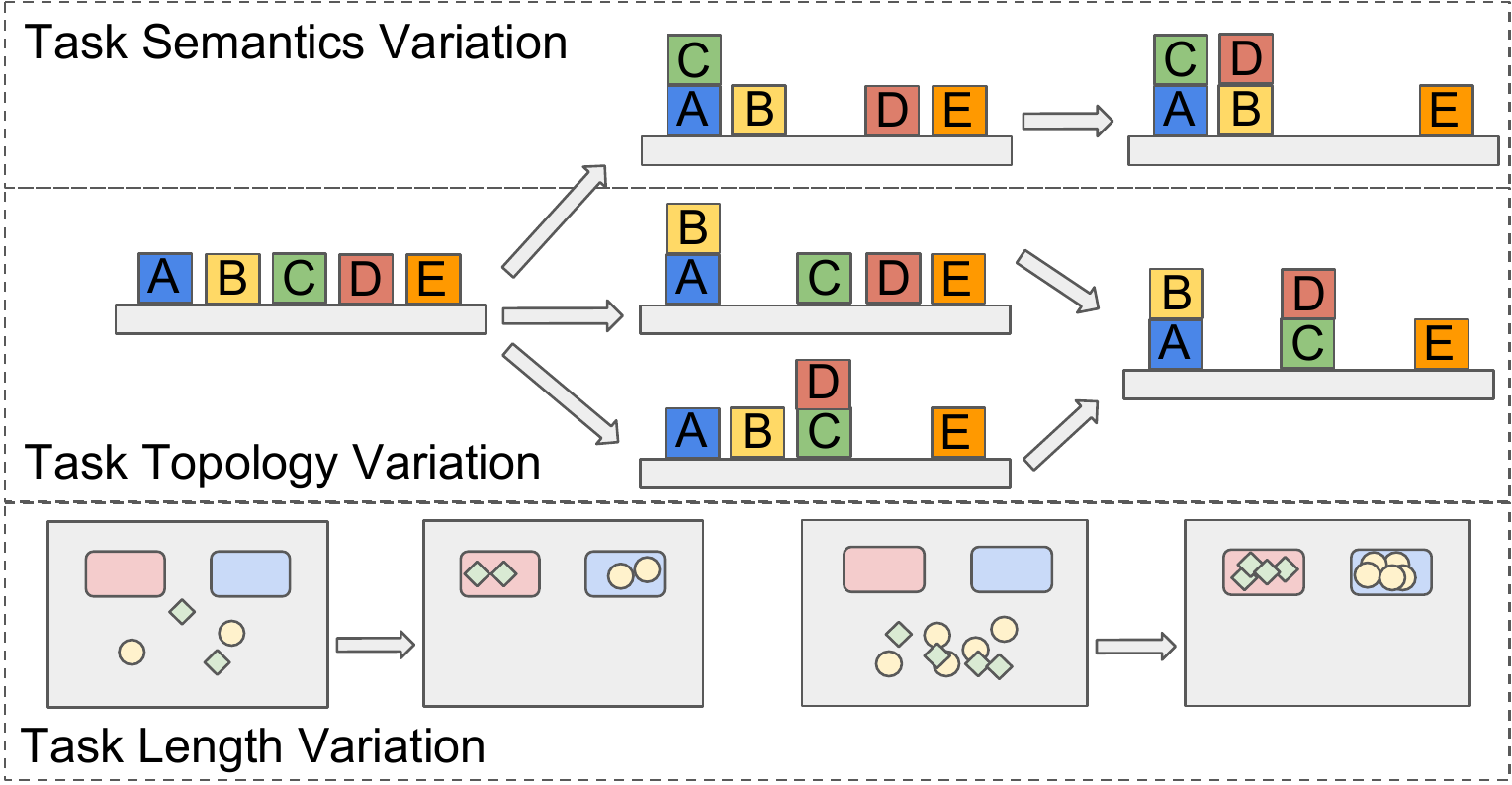}
  \caption{The variability of a task structure consists of changing success conditions (task semantics), variable subtask permutations (task topology), and larger task sizes (task length). We evaluate the ability of our proposed model in generalizing towards these three types of variations.}
  \label{fig:generalization}
  \vspace{-5pt}
\end{figure}

\section{Experimental Setup}
\label{sec:exp}

The goal of our experimental evaluation is to answer the following questions: (1) Does \algoName generalize to changes in all three dimensions of variation: length, topology, and semantics, as illustrated in~\figref{generalization}, (2) Can \algoName use image-based input without access to ground truth state, and (3) Would \algoName also work in real-world tasks which have combinations of these variations.  
We evaluate \algoName in three robot manipulation tasks: Object Sorting, Block Stacking, and Table Clean-up. Each of these tasks requires multiple steps to complete and can be recursively decomposed into repetitive sub-tasks.

\noindent \textbf{Input State Representation}. 
We use an expert policy to generate program execution traces as training data. An expert policy is an agent with hard-coded rules that call programs (\texttt{move\_to}, \texttt{pick\_and\_drop}, etc.) to perform a task. In our experiment, we use the demonstration of a robot carrying out a task as the task specification. 
For all experiments, unless specified, the state representation in the task demonstrations is in the form of object position trajectories relative to the gripper frame.
In the Block Stacking experiment, we also report the results of using a learned object detector to predict object locations and the results of directly using RGB video sequence as state observations and task demonstrations.

\noindent \textbf{Simulator Setup}. We conduct our experiments in a 3D environment simulated using the Bullet Physics engine~\cite{BulletPhysics}.  We use a disembodied PR2 gripper for both gathering training data and evaluation. We also evaluate \algoName on a simulated 7-DoF Sawyer arm with a parallel-jaw gripper as shown in~\figref{pull} and~\figref{cleanup}. Since NTP only considers end-effector pose, the choice of robot does not affect its performance in the simulated environment.

\noindent \textbf{Real-Robot Setup}.
We also demonstrate \algoName's performance on the Block Stacking and the Object Sorting tasks on a 7-DoF Sawyer arm using position control. We use \algoName models that are trained with simulated data. Task demonstration are obtained in the simulator, and the instantiated \algoName models are executed on the robot. All real-robot experiments use object locations relative to the gripper as state observations. A Kinect2 camera is used to localize objects in the 3D scene. 

\noindent \textbf{Evaluation Metrics.} 
We evaluate \algoName on three variations of task structure as illustrated in~\figref{generalization}: 1) \emph{task length}: varying number of steps due to the increasing problem size (e.g., having more objects to transport); 2) \emph{task topology}: variations in permutations of steps of sub-tasks to reach the same end goal (e.g., manipulating objects in different orders); and 3) \emph{task semantics}: the unseen task objectives and success conditions (e.g., placing objects into a different container). 

We evaluate \textit{Task length} on the Object Sorting task varying the number of objects instances from 1 to 10 per category. Further, we evaluate \emph{Task Topology} on the Block Stacking task with different permutations of pick-and-place sub-tasks that lead to the same block configurations. Finally, we evaluate \emph{Task Semantics} on Block Stacking on a held-out set of task demonstrations that lead to unseen block configurations as task objectives.We report success rates for simulation tasks, and we analyze success rates, causes of failure, and proportion of task completed for real-robot evaluation. All objects are randomly placed initially in all of the evaluation tasks in both the simulated and the real-robot setting.

\noindent \textbf{Baselines.}
We compare \algoName to four baselines architecture variations.  
(1) \textbf{Flat} is a non-hierarchical model, similar to \cite{duan2017one}, that takes as input task demonstration and current observation, and directly predicts the primitive APIs instead of calling hierarchical programs. 
(2) \textbf{Flat (GRU)} is the Flat model with a GRU cell.
(3) \textbf{NTP (no scope)} is a variant of the NTP model that feeds the entire demonstration to the subprograms, thereby discarding the scoping constraint.
(4) \textbf{\algoName (GRU)} is a complete \algoName model with a GRU cell. This is to demonstrate that the reactive core network in \algoName can better generalize to longer tasks and recover from unexpected failures due to noise, which is crucial in robot manipulation tasks.

\begin{figure}[!t]
\centering
  \includegraphics[width=0.8\linewidth]{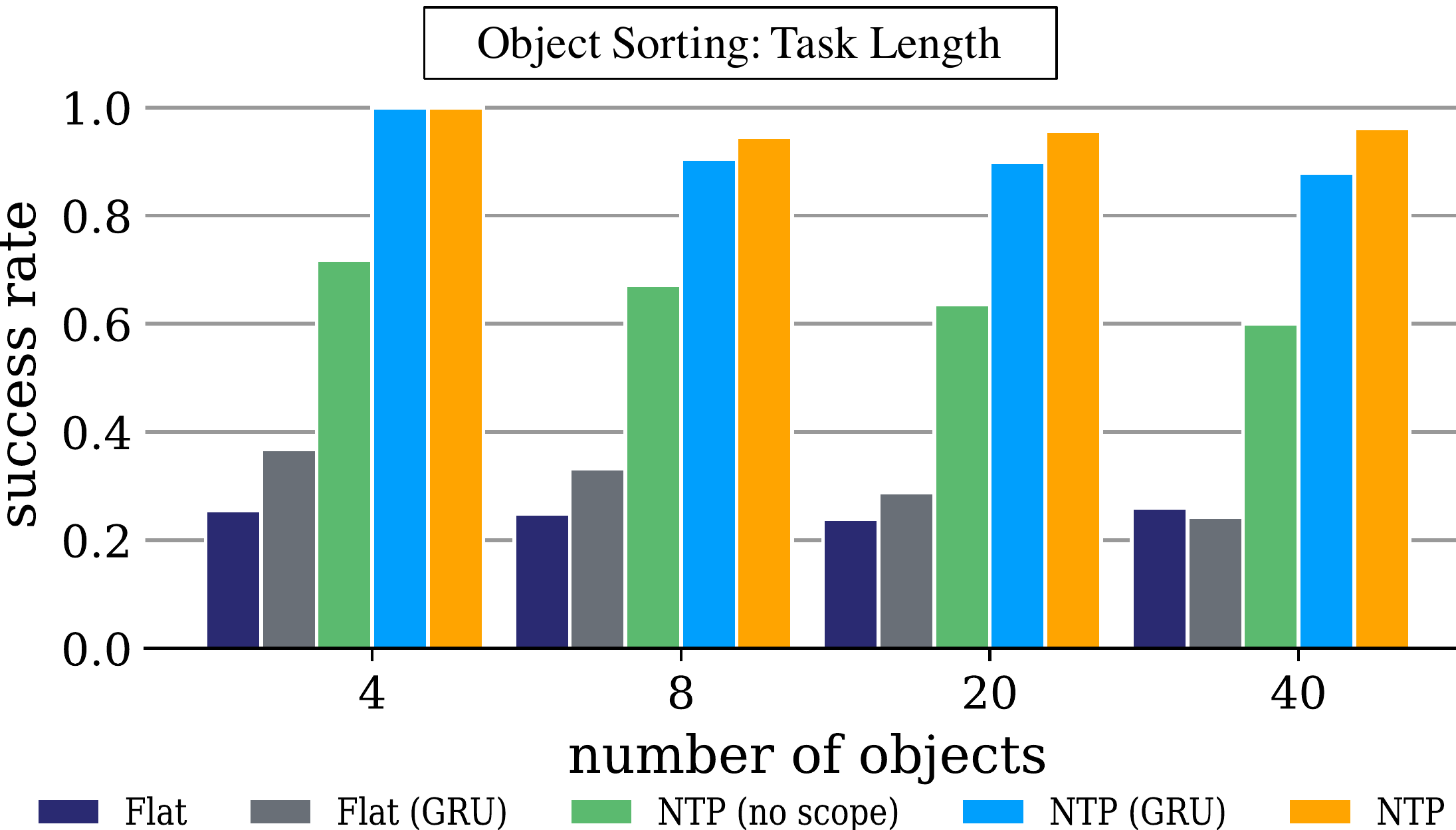}
  \caption{\textbf{Task Length}: Evaluation of the Object Sorting in simulation. The axes represent mean success rate ($y$) with 100 evaluations each and the number of objects in unseen task instances ($x$). \algoName generalizes to increasingly longer tasks while baselines do not.
}
  \label{fig:sorting-exp}
\end{figure}

\begin{figure*}[!t]
\centering
  \includegraphics[width=0.9\linewidth]{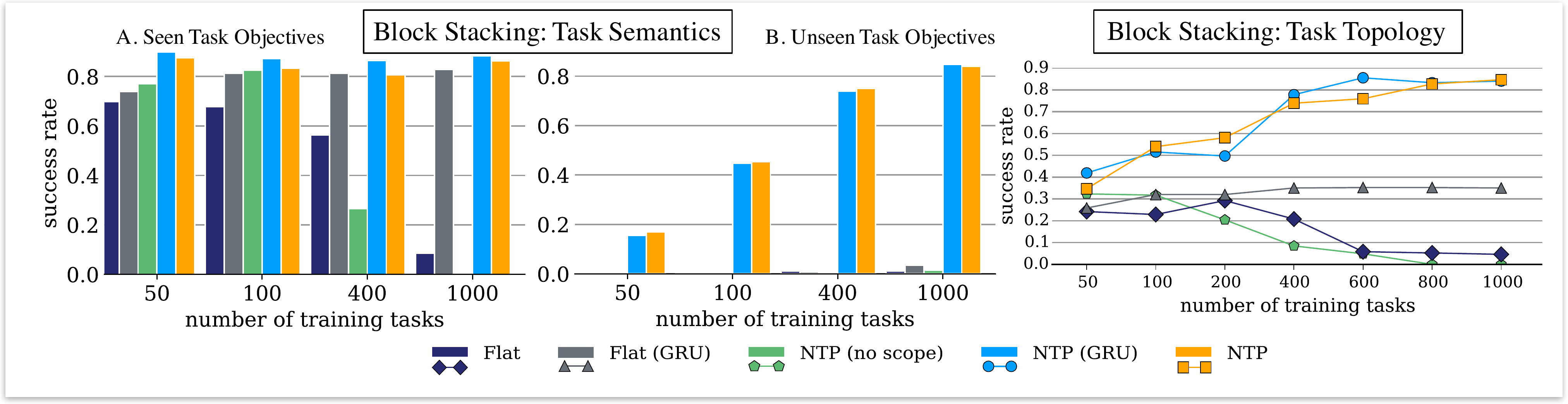}
  \caption{\textbf{Task Semantics}: Simulated evaluation of the Block Stacking. The $x$-axis is the number of tasks used for training. The $y$-axis is the overall success rate. (\textbf{A}) and (\textbf{B}) show that \algoName and its variants generalize better to novel task demonstrations and objectives as the number of training tasks increases.\\  
  \textbf{Task Topology}: Simulated evaluation of the Block Stacking. \algoName shows better performance in task topology generalization as the number of training tasks grows. In contrast, the flat baselines cannot handle topology variability.}
\label{fig:stacking-exp}
\end{figure*}

\begin{figure*}[!t]
\centering

\begin{minipage}[t]{.55\linewidth}
    \centering
    \vspace{-5pt}
    \includegraphics[width=\linewidth]{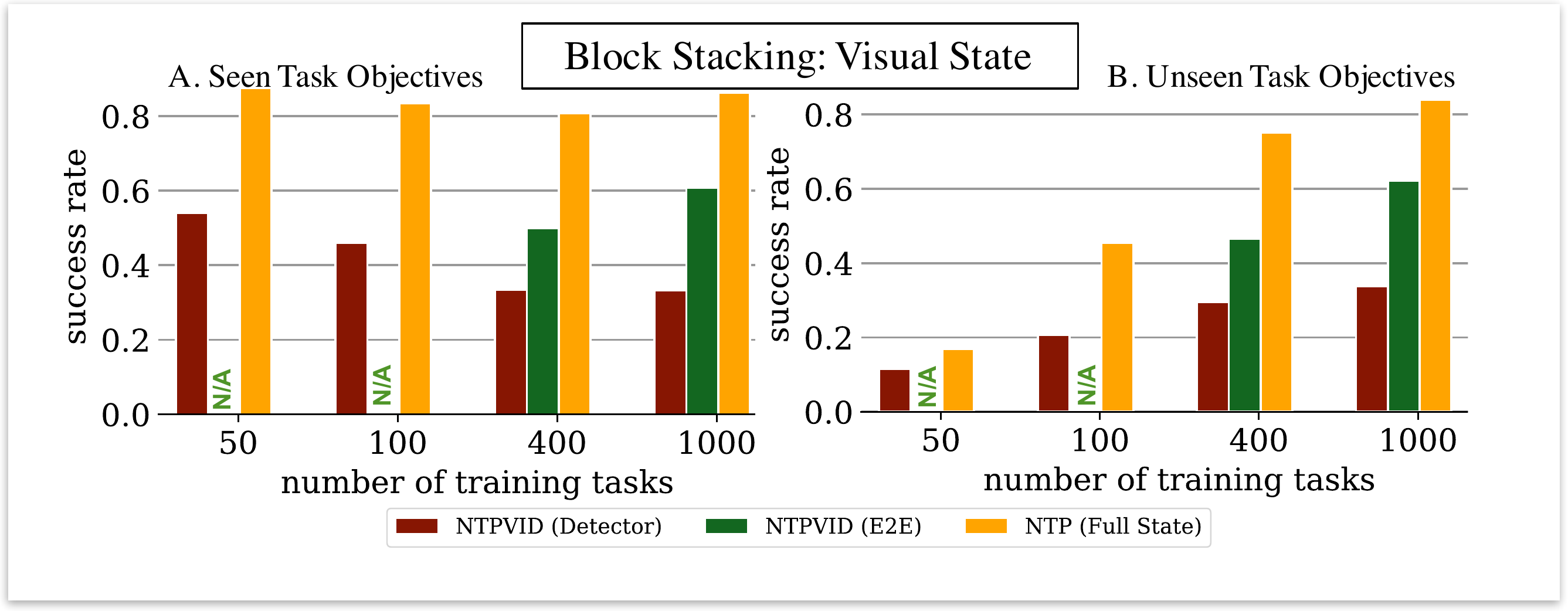}
    \vspace{-5pt}
\end{minipage}%
~
\begin{minipage}[t]{.45\linewidth}
    \centering
    \caption{\textbf{NTP with Visual State:} NTPVID~(Detector) uses an object detector on images which is subsequently used as state in NTP. NTP (E2E) is an end-to-end model trained completely on images with no low-level state information. We note that in the partial observation case (only video), similar learning trends were observed as compared to fully observed case (NTP (Full State)), albeit with a decrease in performance.}
    \label{fig:stacking-vision-exp}
    \vspace{-5pt}
\end{minipage}%
\vspace{-20pt}
\end{figure*}

\section{Experiment 1: Object Sorting}

\noindent\textbf{Setup.} The goal of Object Sorting is to transport objects randomly scattered on a tabletop into their respective shipping containers stated in the task demonstration. We use 4 object categories and 4 containers in evaluating the Object Sorting task. 
In the real robot setup, a toy duck, toy frog, lego block, and marker are used as the objects for sorting, and are sorted into 4 black plastic bins. 
This results in a total of $4^4=256$ category-container combinations (multiple categories may be mapped to the same container). However, as each category can be mapped to 4 possible containers, a minimum of 4 tasks can cover all possible category-container pairs. We select these 4 tasks for training and the other 252 unseen tasks for evaluation. We train all models with 500 trajectories. Each test run is on 100 randomly-selected unseen tasks. 

\noindent\textbf{Simulator.}  As shown in \figref{sorting-exp}, \algoName significantly outperforms the flat baselines. We examine how the task size affects its performance.
We vary the numbers of objects to be transported from 4 to 40 in the experiments. The result shows that \algoName retains a stable and good performance (over 90\%) in longer tasks. On the contrary, the flat models' performances decline from around 40\% to around 25\%, which is close to random. The performance of the NTP (GRU) model also declines faster comparing to the NTP model as the number of objects increases. This comparison illustrates \algoName's ability to generalize towards task length variations. 

\noindent\textbf{Real robot.} \tabref{robot} shows the results of the Object Sorting task on the robot. We use 4 object categories with 3 instances of each category. We carried out a total of 10 evaluation trials on randomly selected unseen Object Sorting tasks. 8 trials completed successfully, and 2 failed due to of manipulation failures: a grasp failure and a collision checking failure.

\section{Experiment 2: Block Stacking}
\noindent\textbf{Setup.} 
The goal of Block Stacking is to stack a set of blocks into a target configuration, similar to the setup in \cite{duan2017one}.
We use 8, 5$\times$5\,cm wooden cubes of different colors both in simulation and with real-robot.
We randomly generate 2000 distinct Block Stacking task instances. Two tasks are 
considered equivalent if they have the same end configuration. 
We use a maximum of 1000 training tasks and 100 trials for each task, leaving the remaining 1000 task instances as unseen test cases. A task is considered successful if the end configuration 
of the blocks matches the task demonstration. 
We evaluate both seen and unseen tasks, i.e., whether the end configuration appears in training set.
We use $N=8$ blocks in our evaluation.

\noindent\textbf{Simulator.} \figref{stacking-exp} shows that all models except the Flat baseline
are able to complete the seen tasks at around 85\% success rate. The performance of the Flat baseline
decreases dramatically when training with more than 400 tasks. It is because the Flat model has very limited expressiveness power to represent complex tasks. The Flat\,(GRU) model performs surprisingly well on the seen tasks. However, as shown in \figref{stacking-exp}, both Flat and Flat\,(GRU) fail to generalize to unseen tasks. We hypothesize that the Flat\,(GRU) baseline simply memorizes the training sequences. On the other hand, \algoName achieves increasingly better performances when the diversity of the training data increases.

We evaluate task topology generalization on random permutations of the pick-and-place sub-tasks that lead to the same end configuration. Specifically, the task variations are generated by randomly shuffling the order that the "block towers" are built in the training tasks. \figref{stacking-exp} illustrates that \algoName generalizes better towards variable topologies when trained on a larger variety of tasks.
We find that increasing the diversity of training data facilitates \algoName to learn better generalizable modules.

Next, we evaluate task semantics generalization. The variability of real-world environments prevents any task-specific policy learning method from training for every possible task.
\figref{stacking-exp}(A) illustrates that \algoName generalizes well to novel task demonstrations and new goals. As the number of training tasks increases,  
both \algoName and its recurrent variation steadily improve their performance on the unseen tasks. When trained with 1000 tasks, their performances on unseen tasks are almost on par with that of seen tasks.

The performance gaps between \algoName (no\,scope) and \algoName highlight the benefit of the scoping constraint. \algoName (no\,scope) performance drops gradually as the task size grows implying that the programs in \algoName learn modularized and reusable semantics due to information hiding, which is crucial to achieving generalization towards new tasks.

\noindent\textbf{Real robot.} \tabref{robot} shows the results of the Object Sorting task in the real world setting. We carried out 20 trails of randomly selected unseen Block Stacking tasks. Out of the 2 failure cases, one is caused by an incorrect placing; the other is caused by the gripper knocking down a stacked tower and not able to recover from the error.

\vspace{-5pt}
\subsection{Adversarial Dynamics}
We show that the reactive core network in \algoName enables it to better recover from failures compared to its recurrent variation. We demonstrate this by performing Block Stacking under an adversary. Upon stacking each block, an adversary applies a force to the towers with a probability of 25\%. The force can knock down the towers. We evaluate \algoName and its recurrent variant on the 1000 unseen tasks. Table~\ref{fig:recover} shows that under the same adversary, the success rate of \algoName with the GRU core decreases by 46\%, whereas the success rate of \algoName only decreases by 20\%. This indicates that a reactive model is more robust against unexpected failures as its behavior is less history dependent than the recurrent counterpart. We also demonstrate this feature in the supplementary video in the real world setting.

\vspace{-5pt}
\subsection{\algoName with Visual State}
This experiment examines the ability of \algoName to learn when demonstrations come in the form of videos and the state is a single image. Unlike the full state information used in experiments thus far, we train an \algoName model NTPVID\,(E2E) to jointly learn a policy and task-relevant features without explicit auxiliary supervision.
An alternative is to use a 2-phase pipeline with an object detector as state preprocessor for \algoName, termed as NTPVID\,(Detector). The detector is a separately trained CNN to predict object position in $\mathbb{R}^3$.

We explore these results in \figref{stacking-vision-exp}, where we see compare the visual models (NTPVID\,(E2E) and NTPVID\,(Detector)) against the best full state model (NTP), all trained on 100 demonstrations per task, for a varying number of tasks. For NTPVID (E2E) we use a 7-layer convolutional network, which takes as input a $64 \times 64$ image and outputs a length 128 feature vector. For NTPVID (Detector) , we use a VGG16 based architecture, predicting the position of the $N$-task objects from an input image of size $224 \times 224$. 

We note that NTPVID (E2E) outperforms NTPVID (Detector) and achieves a higher success rate despite only having partial state information. Both of these methods are inferior to the full-state \algoName version.
NTPVID (Detector) does not generalize due to task-specific state representation, and cascading errors in detection propagate to \algoName reducing performance even when using a very deep network for the detection. The detector errors are Gaussian with standard deviation of 2 cm.
However, this performance comes at a computational cost.
NTPVID (E2E) was trained on 1000 training tasks for 10 days on 8 Nvidia Titan X GPUs. NTPVID (Detector) was trained for 24 hours on a single GPU.  Due to computational cost, we only evaluated NTPVID (E2E) on 400 and 1000 training tasks.

\section{Experiment 3: Table Clean-up}

\noindent\textbf{Setup.} We also evaluate NTP on the Table Clean-up task, which exemplifies a practical real-world task. Specifically, the goal of the task is to clear up to 4 white plastic bowls and 20 red plastic forks into a bin such that the resulting stack of bowls and forks can be steadily carried away in a tray. Task variation comes in task length, where the number of utensils varies, and task topology, where the ordering in which bowls are stacked can vary. Using trajectories as demonstrations and object positions as state space, a model is trained using 1000 task instances. 

\noindent{\textbf{Simulator.}}
We observe that performance varies between 55\%-100\% where increasing errors with more objects  are attributed to failures in collision checking, not incorrect decisions from \algoName. The result shows that \algoName retains its generalization ability in a task that requires multiple dimensions of generalization.

\noindent{\textbf{Real robot.}}
We have also transferred the trained model on the real-Sawyer arm to evaluate the feasibility as shown in \figref{cleanup}. We demonstrate this task in the supplementary video in the real world setting.

\begin{table}[!t]
    \centering
  \caption{\textbf{Real Robot Evaluation}: Results of 20 unseen Block Stacking evaluations and 10 unseen sorting evaluations on Sawyer robot for the NTP model trained on simulator. NTP Fail denotes an algorithmic mistake, while Manip. Fail denotes a mistake in physical interaction (e.g. grasping failures and collisions).}
  \label{tab:robot}
  \resizebox{0.8\linewidth}{!}{%
    \begin{tabular}{lcccccc}
    \hline
    \rowcolor[HTML]{CBCEFB}
    \textbf{Tasks}   & \# Trials  & Success & \algoName Fail & Manip. Fail \\ \hline 
    \cellcolor[HTML]{FFCC67}
    Blk. Stk.  & 20 & 0.9     & 0.05 & 0.05  \\
    \rowcolor[HTML]{EFEFEF}
    \cellcolor[HTML]{FFCC67}
    Sorting   & 10 & 0.8     & 0 & 0.20  \\ \hline
  \end{tabular}
  }
\end{table}

\begin{table}[!t]
    \centering
    \vspace{-10pt}
  \caption{\textbf{Adversarial Dynamics}: Evaluation results of the Block Stacking Task in a simulated adversarial environment. We find that NTP with GRU performs markedly worse with intermittent failures.}
  \label{fig:recover}
    \begin{tabular}{lcc}
    \hline
    \rowcolor[HTML]{CBCEFB}
    \textbf{Model}     & \textbf{No failure} & \textbf{With failures} \\ \hline
    \cellcolor[HTML]{FFCC67}\algoName      & 0.863      & 0.663   \\
    \rowcolor[HTML]{EFEFEF}
    \cellcolor[HTML]{FFCC67}\algoName (GRU) & 0.884      & 0.422   \\ \hline
  \end{tabular}
  \vspace{-5pt}
\end{table}

\section{Discussion \& Future Work}
We introduced Neural Task Programming (NTP), a meta-learning framework that learns modular and reusable neural programs for hierarchical tasks. We demonstrate NTP's strengths in three robot manipulation tasks that require prolonged and complex interactions with the environment. NTP achieves generalization towards task length, topology, and semantics. This work opens up the opportunity to use generalizable neural programs for modeling hierarchical tasks. For future work, we intend to 1) improve the state encoder to extract more task-salient information such as object relationships, 2) devise a richer set of APIs such as velocity and torque-based controllers, and 3) extend this framework to tackle more complex tasks on real robots.

\begin{figure}[t]
\centering
  \includegraphics[width=0.9\linewidth]{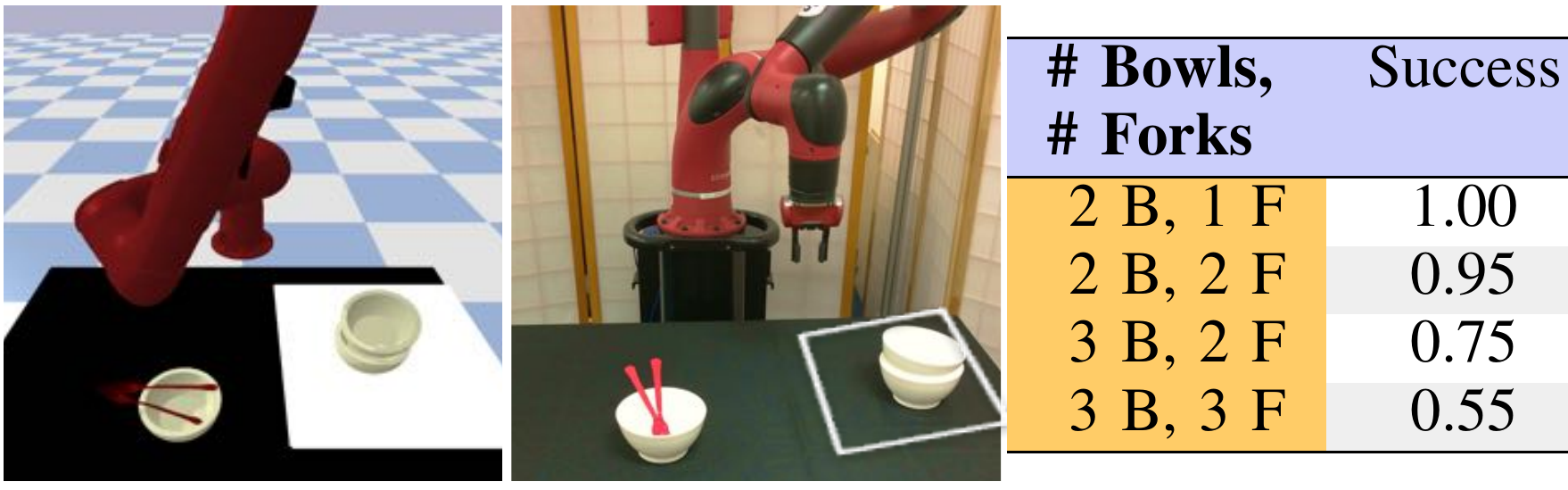}
  \vspace{-5pt}
  \caption{\textbf{Table Clean-up}:  in simulated and real environment.The table shows success rates for varying numbers of forks and bowls in the simulated evaluation.}
  \label{fig:cleanup}
  \vspace{-10pt}
\end{figure}

{\footnotesize 
\section*{Acknowledgment}
This research was performed at the SVL at Stanford in
affiliation with the Stanford AI Lab, Stanford-Toyota AI Center. 
}

\renewcommand*{\bibfont}{\footnotesize}
\begin{flushright}
\printbibliography 
\end{flushright}

\end{document}